\documentclass[conference]{IEEEtran}
\IEEEoverridecommandlockouts

\usepackage{cite}
\usepackage{amsmath,amssymb,amsfonts}
\usepackage{algorithmic}
\usepackage{graphicx}
\usepackage{amsmath}
\usepackage{amssymb}
\usepackage{multirow}
\usepackage{array}
\def\BibTeX{{\rm B\kern-.05em{\sc i\kern-.025em b}\kern-.08em
    T\kern-.1667em\lower.7ex\hbox{E}\kern-.125emX}}
\begin{document}

\title{Human-Guided Reasoning with Large Language Models for Vietnamese Speech Emotion Recognition\\

}
\author{
\IEEEauthorblockN{Truc Nguyen$^{1,2}$, Then Tran$^{1,2}$, Binh Truong$^{1,2}$, and Phuoc Nguyen T.~H.$^{1,2*}$}
\IEEEauthorblockA{$^{1}$University of Information Technology, Ho Chi Minh City, Vietnam\\
$^{2}$Vietnam National University Ho Chi Minh City, Ho Chi Minh City, Vietnam}\\[2pt]
\texttt{\small 23521670@gm.uit.edu.vn, 23521471@gm.uit.edu.vn, binhtc.18@grad.uit.edu.vn, phuocnth@uit.edu.vn}
}
\maketitle

\begin{abstract}
Vietnamese Speech Emotion Recognition (SER) remains challenging due to ambiguous acoustic patterns and the lack of reliable annotated data, especially in real-world conditions where emotional boundaries are not clearly separable. To address this problem, this paper proposes a human–machine collaborative framework that integrates human knowledge into the learning process rather than relying solely on data-driven models. The proposed framework is centered around LLM-based reasoning, where acoustic feature-based models are used to provide auxiliary signals such as confidence and feature-level evidence. A confidence-based routing mechanism is introduced to distinguish between easy and ambiguous samples, allowing uncertain cases to be delegated to LLMs for deeper reasoning guided by structured rules derived from human annotation behavior. In addition, an iterative refinement strategy is employed to continuously improve system performance through error analysis and rule updates. Experiments are conducted on a Vietnamese speech dataset of 2,764 samples across three emotion classes (calm, angry, panic), with high inter-annotator agreement (Fleiss’ Kappa = 0.8574), ensuring reliable ground truth. The proposed method achieves strong performance, reaching up to \textbf{86.59\% accuracy} and \textbf{Macro F1 around 0.85–0.86}, demonstrating its effectiveness in handling ambiguous and hard-to-classify cases. Overall, this work highlights the importance of combining data-driven models with human reasoning, providing a robust and model-agnostic approach for speech emotion recognition in low-resource settings.
\end{abstract}


\section{Introduction}
Speech Emotion Recognition (SER) is a key task in human–computer interaction, with increasing importance in healthcare applications such as telehealth monitoring, mental health assessment, and emergency detection systems \cite{b1,b2}. By analyzing emotional cues in speech, SER systems can help detect critical states like stress, anxiety, or panic for timely intervention. However, for low-resource languages such as Vietnamese, SER remains challenging due to limited standardized datasets and complex acoustic characteristics influenced by tone, region, and speaking style.

A critical issue in Vietnamese SER is the ambiguity of acoustic features across emotion classes, particularly in medically relevant states. Emotions such as \textit{angry} and \textit{panic} often exhibit overlapping patterns in pitch and energy, making them difficult to distinguish reliably. This becomes especially problematic in healthcare scenarios, where misclassification of panic may lead to missed early warnings. In addition, emotion annotation is inherently subjective, introducing uncertainty even in ground truth labels \cite{b3,b8}.

Existing approaches are primarily data-driven, relying on machine or deep learning models trained on acoustic features such as MFCC, pitch, and energy \cite{b1,b4,b5}. While effective in controlled settings, they often fail in ambiguous cases common in real-world scenarios \cite{b6,b7}. Moreover, these methods lack the ability to incorporate human reasoning, which is crucial for interpreting emotional signals \cite{b3,b8}. This highlights a key limitation: current SER systems cannot effectively combine acoustic patterns with human-like reasoning to resolve ambiguity in practical applications.

Motivated by these challenges, we argue that human knowledge plays a key role in addressing ambiguity, especially in healthcare contexts where interpretability and reliability are critical \cite{b8}. Human annotators rely on implicit reasoning patterns, such as linking rapid pitch variations and unstable energy to panic-like states \cite{b3}. These patterns are not captured by conventional models but are essential for accurate recognition in ambiguous scenarios.

To address this, we propose a human--machine collaborative framework that integrates machine learning with reasoning. The system first performs classification using an SVM based on acoustic features \cite{b1}, followed by a confidence-based routing mechanism to identify uncertain samples. These cases are then refined using LLM-based reasoning \cite{b9}, guided by rules derived from human annotation behavior. This hybrid design effectively combines data-driven learning with human-like reasoning, improving both interpretability and reliability.

The main contributions of this work are as follows:

\begin{itemize}
    \item We construct a high-quality Vietnamese speech emotion dataset with 2,764 samples collected from diverse real-world sources, covering three emotion classes (calm, angry, panic). The dataset is carefully annotated by multiple annotators with strong inter-annotator agreement (Fleiss’ Kappa = 0.8574), and is designed to capture ambiguous and clinically relevant emotional states.

    \item We introduce a novel human--machine collaborative \textbf{methodology} for speech emotion annotation, where human reasoning is explicitly formalized and integrated into LLM-based inference.

    \item We develop a method to transform implicit human annotation knowledge into explicit reasoning rules, enabling the system to incorporate human-like decision-making for improved interpretability and robustness.
\end{itemize}
This approach not only improves performance in ambiguous cases but also enhances interpretability and reliability, making it particularly suitable for real-world healthcare applications.

\section{Related Work}

Speech Emotion Recognition (SER) has been widely studied, evolving from handcrafted features to deep learning-based representations \cite{b1,b7}.

\textbf{Feature-based and traditional approaches.}  
Early SER systems rely on handcrafted acoustic features such as pitch, energy, and MFCC \cite{b1}. Benchmark efforts like the INTERSPEECH 2011 Speaker State Challenge \cite{b2} demonstrate their effectiveness. Tools such as OpenSMILE \cite{b4} and librosa \cite{b5} are widely used, followed by classifiers such as SVM. However, these approaches depend heavily on feature engineering and often fail to generalize in real-world and low-resource settings \cite{b6}.

\textbf{Deep learning-based approaches.}  
Recent work adopts deep neural networks for end-to-end learning, including CNNs and RNNs \cite{b11,b12,b13}, with multimodal extensions further improving performance \cite{b9}. Despite these advances, such models remain purely data-driven and struggle with ambiguous emotional boundaries in practical scenarios \cite{b7}.

\textbf{Self-supervised representation learning.}  
Methods such as wav2vec 2.0 \cite{b14} leverage large-scale unlabeled data to learn robust speech representations. While effective, these approaches still lack mechanisms to incorporate human reasoning and interpretability.

\textbf{Human-in-the-loop learning.}  
Human-in-the-loop approaches incorporate human knowledge into machine learning processes \cite{b3,b8}, improving interpretability in critical domains such as healthcare. However, they are not specifically designed for SER and typically do not integrate acoustic modeling with structured reasoning.

\textbf{LLM-based reasoning.}  
Recent advances in large language models demonstrate strong capabilities in structured reasoning and knowledge integration. However, their application to SER remains limited, particularly in combining acoustic signals with human-derived reasoning.

\textbf{Vietnamese speech emotion recognition.}  
Research on Vietnamese SER is still limited. Existing work such as VNEMOS \cite{b15} adopts deep learning but remains purely data-driven and does not address ambiguity through reasoning.

\textbf{Research gap.}  
Existing methods fail to jointly model acoustic patterns with structured human-like reasoning, which is critical for resolving ambiguity in real-world and high-risk scenarios.

\textbf{Our approach.}  
To address this gap, we propose a human--machine collaborative methodology centered on LLM-based reasoning, where machine learning provides auxiliary evidence and human knowledge is explicitly formalized into reasoning rules for more reliable and interpretable emotion recognition.

\section{Proposed System}

\subsection{System Overview}

\begin{figure*}[t]
\centering
\includegraphics[width=0.95\textwidth]{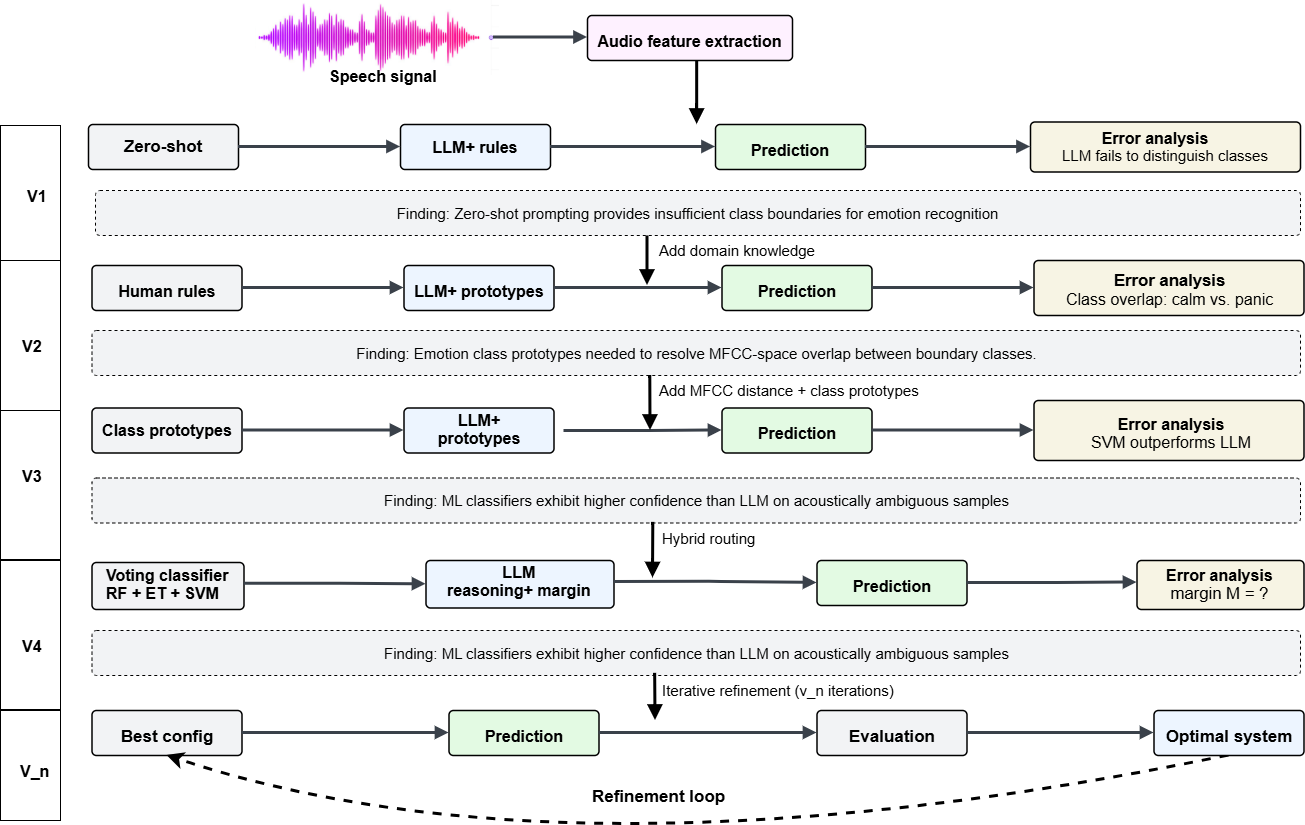}
\caption{Overall architecture of the proposed human--machine collaborative framework.}
\label{fig:pipeline}
\end{figure*}
Figure~\ref{fig:pipeline} illustrates the overall architecture of the proposed system. 
The input is a raw speech waveform, from which acoustic features (e.g., pitch, energy, and MFCC) are extracted to represent emotional characteristics.

The system consists of three main components: (1) acoustic feature representation, (2) an LLM-based reasoning framework, and (3) a hybrid inference mechanism. The reasoning framework incorporates human-derived rules to infer emotion labels, while the hybrid mechanism leverages machine learning signals to handle confident cases.

In the data flow, speech is first converted into acoustic features and processed by the LLM reasoning module. Ambiguous samples are further refined through hybrid inference by combining LLM reasoning with machine learning outputs.

The system produces a final emotion label for each input, followed by an iterative refinement loop that updates reasoning strategies based on prediction errors.

\subsection{Feature Representation}
The input to the system is a raw speech waveform, which is transformed into a structured acoustic representation. Three types of features are extracted to capture emotional characteristics: pitch, energy, and Mel-Frequency Cepstral Coefficients (MFCC) \cite{b1,b4}.

Pitch reflects tonal variations related to emotional intensity, while energy captures loudness patterns associated with arousal. MFCC provides a compact representation of the spectral structure of speech, useful for distinguishing emotional states.

These features are computed at the frame level and aggregated into a fixed-length vector, which serves as input to the subsequent reasoning framework.

\subsection{Reasoning Framework}

The proposed system employs an LLM-based reasoning framework that performs structured inference over acoustic representations instead of direct classification.

Given the extracted features, acoustic cues are first transformed into a structured description, which is provided to the LLM together with human-derived rules and known confusion patterns between emotion classes. This enables the model to perform reasoning by jointly considering feature trends (e.g., stability, variability) and class-specific characteristics.

Human knowledge is explicitly encoded as heuristic rules (e.g., high variance in pitch and energy indicates \textit{panic}, while stable patterns correspond to \textit{calm}), guiding the reasoning process in ambiguous cases.

By integrating structured feature descriptions with rule-based reasoning, the framework enhances interpretability and improves decision consistency when acoustic evidence alone is insufficient.

\subsection{Hybrid Inference}

The LLM serves as the primary reasoning engine of the system, responsible for making the final decision based on structured feature descriptions and human-derived rules. 

Machine learning outputs are incorporated as auxiliary signals that provide confidence estimates and support the reasoning process. 

A confidence-aware routing strategy is used to determine when additional reasoning is required. For samples with high confidence, the decision can be made directly, while ambiguous cases are further analyzed by the LLM through deeper reasoning.

This design emphasizes reasoning-driven inference, where machine learning acts as supportive evidence rather than the main decision-making component.

\subsection{Iterative Refinement}

The proposed system employs an iterative refinement process driven by error analysis.

Misclassified samples are analyzed to identify recurring error patterns and ambiguities. These patterns are then translated into updated reasoning rules and prompt adjustments, forming a feedback loop that progressively improves decision consistency.

This mechanism enables the framework to continuously incorporate new knowledge, enhancing robustness, particularly in resolving ambiguous emotion classes.

\section{Dataset}

\subsection{Dataset Collection}

In this study, we construct a Vietnamese speech emotion dataset covering the three major dialect regions (Northern, Central, and Southern), aiming to capture variability in emotional expression in real-world scenarios, including healthcare contexts.

\textbf{Data Sources.}
The dataset consists of \textbf{2,764 audio segments} collected from \textbf{28 sources}, including 5 movies, 10 entertainment programs, and 13 interview-based programs.

\textbf{Dataset Scale.}
The total duration is \textbf{14,841.79 seconds} (\textbf{247.36 minutes}, \textbf{4.12 hours}), with a storage size of approximately \textbf{453.09 MB}, which is sufficient for meaningful training and evaluation.

\textbf{Data Diversity.}
Multiple sources are used to capture diverse emotional expressions across different communication contexts. This design emphasizes variability in how emotions are expressed, improving generalization beyond controlled settings.

\textbf{Class Distribution.}
The dataset includes three emotion classes: \textit{angry} (\textbf{942}, 34.1\%), \textit{calm} (\textbf{980}, 35.5\%), and \textit{panic} (\textbf{842}, 30.5\%). 
This relatively balanced distribution reflects real-world conditions and introduces additional challenges for emotion recognition.

\begin{figure}[htbp]
\centering
\includegraphics[width=\linewidth]{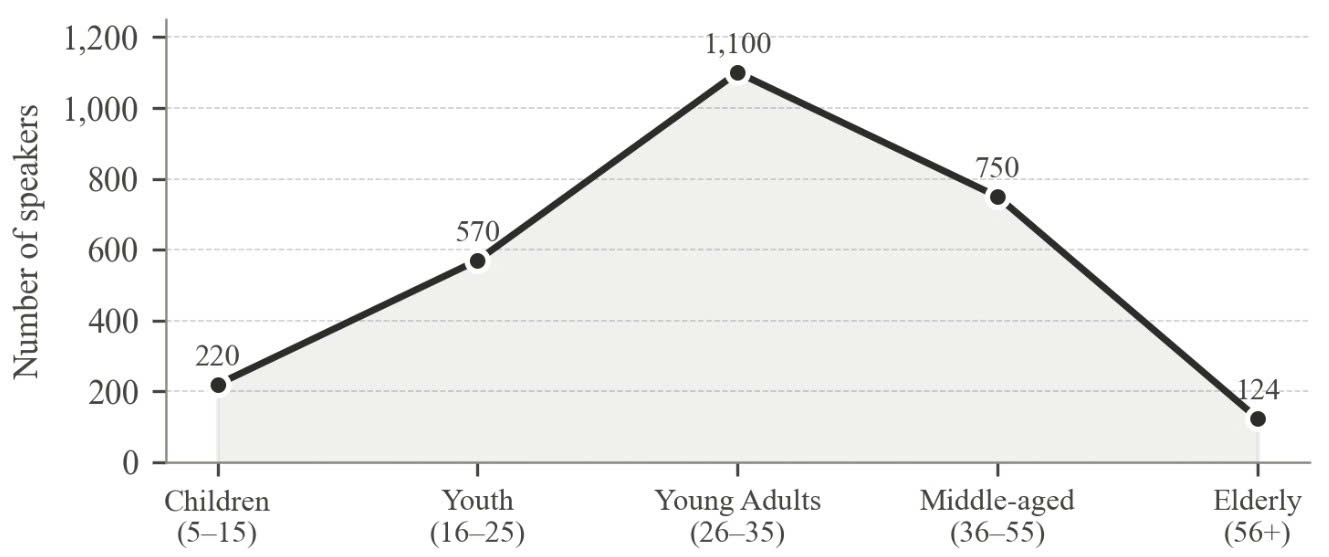}
\caption{Distribution of speakers across different age groups in the proposed dataset.}
\label{fig:age_distribution}
\end{figure}

Figure~\ref{fig:age_distribution} shows the distribution of speakers across age groups. The dataset maintains a balanced representation, supporting better generalization to real-world users.

\textbf{Recording Conditions.}
Recordings are collected in real-world environments, including healthcare-related scenarios, with background noise, overlapping speech, and device variability. While these factors increase complexity, they enhance the realism of the dataset.

\textbf{Dataset Characteristics.}
Overall, the dataset captures realistic emotional expressions in Vietnamese speech, making it suitable for developing and evaluating robust speech emotion recognition systems.

\subsection{Data Preprocessing and Annotation}

To ensure data quality and annotation reliability, we design a unified pipeline that integrates preprocessing and human labeling.

\textbf{Audio Preprocessing.}
All recordings are standardized (16 kHz, mono, amplitude normalization). Noise is reduced using automatic methods and manual inspection. Silence is removed via VAD, and long recordings are segmented into shorter utterances with manual correction. Low-quality samples are filtered out.

\textbf{Annotation Process.}
Each segment is labeled by three annotators using a three-class scheme: \textit{angry}, \textit{calm}, and \textit{panic}, following guidelines based on acoustic and contextual cues such as pitch, intensity, and speaking rate.

\textbf{Emotion Definition.}
\textit{Calm} exhibits stable pitch and low energy variation, \textit{angry} shows higher intensity and pitch, and \textit{panic} is characterized by high variability in pitch, energy, and speaking rate.

\textbf{Quality Control.}
Disagreements are resolved through re-evaluation or additional annotation, and ambiguous samples are refined or removed.

This pipeline ensures high-quality audio data and reliable annotations for subsequent modeling and evaluation.

\subsection{Inter-Annotator Agreement}

To assess annotation reliability, we use both \textit{Fleiss’ Kappa} and \textit{Cohen’s Kappa}. Fleiss’ Kappa measures overall agreement among the three annotators, while Cohen’s Kappa evaluates pairwise consistency.

\begin{equation}
\kappa = \frac{\bar{P} - \bar{P_e}}{1 - \bar{P_e}} \quad (1)
\qquad
\kappa = \frac{p_o - p_e}{1 - p_e} \quad (2)
\end{equation}

where $\bar{P}$ and $\bar{P_e}$ denote the observed and expected agreement across multiple annotators, while $p_o$ and $p_e$ represent the observed and expected agreement between two annotators.

\begin{table}[h]
\centering
\caption{Inter-annotator agreement (Kappa scores)}
\label{tab:kappa}
\renewcommand{\arraystretch}{1.2}
\setlength{\tabcolsep}{6pt}
\small
\begin{tabular}{lccccc}
\hline
\textbf{Metric} 
& \textbf{Fleiss $\kappa$} 
& \textbf{Avg $\kappa$} 
& \textbf{A--B} 
& \textbf{A--C} 
& \textbf{B--C} \\
\hline
\textbf{Value} 
& 0.8574 
& 0.8575 
& 0.8573 
& 0.8394 
& 0.8757 \\
\hline
\end{tabular}
\end{table}

\textbf{Results.}
As shown in Table~\ref{tab:kappa}, the dataset achieves a high level of agreement, with a Fleiss’ Kappa score of \textbf{0.8574}. 
Pairwise Cohen’s Kappa values range from \textbf{0.8394} to \textbf{0.8757}, indicating strong consistency across annotators.

According to standard interpretation guidelines, these scores correspond to \textit{substantial to almost perfect agreement}, demonstrating the reliability of the annotation process.

\begin{table}[h]
\centering
\caption{Annotator-wise accuracy results}
\label{tab:annotator_acc}
\small
\begin{tabular}{l c c c c}
\hline
\textbf{Annotator} & \textbf{Overall} & \textbf{Angry} & \textbf{Calm} & \textbf{Panic} \\
\hline
Annotator A & 87.8\% & 90.0\% & 90.6\% & 91.3\% \\
Annotator B & 90.5\% & 91.0\% & 91.6\% & 89.1\% \\
Annotator C & 90.4\% & 91.1\% & 89.7\% & 90.8\% \\
\hline
\end{tabular}
\end{table}
\section{Experimental Results}

\subsection{Experimental Setup}

To evaluate the effectiveness of the proposed approach, we conduct experiments on the constructed Vietnamese speech emotion dataset.

\textbf{Data Split.}
The dataset is divided into four subsets: \textit{Set1} (706 samples), \textit{Set2} (691 samples), \textit{Set3} (696 samples), and \textit{Test} (671 samples).  The three sets (\textit{Set1--Set3}) are used for training and validation under different configurations, while the \textit{Test} set is reserved for final evaluation. This split ensures fair assessment and allows analysis of model consistency across different data partitions.

\textbf{Proposed Method.}
Our approach adopts a hybrid framework that combines data-driven signals with a reasoning-based module. Initial predictions are generated from acoustic features, while ambiguous samples are further refined through structured reasoning to improve classification performance.

\textbf{Implementation Details.}
All experiments are implemented in Python using standard libraries. The models are trained and evaluated under the same conditions to ensure fair comparison.

\subsection{Evaluation Metrics}

To evaluate model performance, we use standard classification metrics, including accuracy, precision, recall, and F1-score.

\textbf{Accuracy} measures the overall proportion of correctly classified samples.

\textbf{Precision}, \textbf{Recall}, and \textbf{F1-score} are computed to provide a more detailed evaluation, especially in the presence of class imbalance.

The F1-score is defined as:

\begin{equation}
F1 = \frac{2 \cdot Precision \cdot Recall}{Precision + Recall}
\end{equation}

In addition, we report \textbf{Macro F1-score} to equally weight all classes, ensuring that performance on minority classes is properly evaluated.

\subsection{Results and Discussion}
\begin{table}[h]
\centering
\caption{PERFORMANCE OF THE PROPOSED SYSTEM ACROSS THREE INDEPENDENT SPLITS USING QWEN2.5-7B}
\label{tab:results}
\renewcommand{\arraystretch}{1.1}
\setlength{\tabcolsep}{4pt}
\small
\begin{tabular}{l l c c c c}
\hline
\textbf{Split} & \textbf{Version} & \textbf{Acc (\%)} & \textbf{Prec} & \textbf{Rec} & \textbf{F1} \\
\hline
\multirow{5}{*}{Set 1}
& v1\_basic   & 37.54 & 0.46 & 0.33 & 0.185 \\
& v2\_rules   & 45.47 & 0.73 & 0.48 & 0.416 \\
& v3\_refined & 54.39 & 0.74 & 0.56 & 0.533 \\
& v4\_hybrid  & \textbf{85.13} & \textbf{0.85} & \textbf{0.85} & \textbf{0.847} \\
& v5\_auto    & 34.70 & 0.24 & 0.36 & 0.273 \\
\hline
\multirow{5}{*}{Set 2}
& v1\_basic   & 32.85 & 0.24 & 0.33 & 0.169 \\
& v2\_rules   & 43.27 & 0.71 & 0.47 & 0.387 \\
& v3\_refined & 53.98 & 0.73 & 0.57 & 0.534 \\
& v4\_hybrid  & \textbf{86.11} & \textbf{0.86} & \textbf{0.86} & \textbf{0.858} \\
& v5\_auto    & 37.63 & 0.13 & 0.33 & 0.182 \\
\hline
\multirow{5}{*}{Set 3}
& v1\_basic   & 31.75 & 0.37 & 0.34 & 0.170 \\
& v2\_rules   & 48.13 & 0.74 & 0.48 & 0.424 \\
& v3\_refined & 55.60 & 0.70 & 0.56 & 0.539 \\
& v4\_hybrid  & \textbf{84.63} & \textbf{0.85} & \textbf{0.85} & \textbf{0.847} \\
& v5\_auto    & 31.18 & 0.10 & 0.33 & 0.158 \\
\hline
\end{tabular}
\end{table}
Table~\ref{tab:results} shows the performance of the pipeline on three data splits (set1, set2, set3) using the Qwen2.5-7B model.

The results follow a clear pattern. The initial version (v1) performs poorly (31–37\% accuracy), indicating that naive LLM-based prediction without structured guidance is insufficient. Performance improves in v2 and v3 after incorporating human-derived rules and refining the reasoning process, demonstrating the effectiveness of structured knowledge in guiding inference.

The best results are achieved in v4 (hybrid), reaching \textbf{84–86\% accuracy}. This significant improvement can be attributed to the complementary strengths of the two components: the machine learning model effectively handles samples with clear acoustic patterns, while the LLM resolves ambiguous cases through rule-based reasoning. This combination enables more robust handling of overlapping emotion classes.

In particular, the improvement is most evident in ambiguous cases such as \textit{angry} versus \textit{panic}, where acoustic features overlap and require reasoning beyond statistical patterns.

In contrast, v5 shows a clear performance drop, indicating that uncontrolled or unguided reasoning may introduce noise and reduce consistency. This highlights the importance of structured reasoning and controlled integration of LLM outputs in the proposed framework.

\begin{table}[h]
\centering
\caption{Performance across different LLM backbones}
\label{tab:backbone}
\renewcommand{\arraystretch}{1.1}
\setlength{\tabcolsep}{4pt}
\small
\begin{tabular}{l l c c c c}
\hline
\textbf{Model} & \textbf{Version} & \textbf{Acc} & \textbf{Prec} & \textbf{Rec} & \textbf{F1} \\
\hline
\multirow{5}{*}{Qwen2.5-7B}
& v1 & 34.58 & 0.12 & 0.33 & 0.17 \\
& v2 & 43.52 & 0.70 & 0.47 & 0.40 \\
& v3 & 53.80 & 0.72 & 0.57 & 0.54 \\
& v4 & \textbf{85.84} & \textbf{0.85} & \textbf{0.85} & \textbf{0.85} \\
& v5 & 58.57 & 0.57 & 0.57 & 0.57 \\
\hline
\multirow{5}{*}{Qwen2.5-14B}
& v1 & 36.66 & 0.12 & 0.33 & 0.18 \\
& v2 & 49.03 & 0.77 & 0.52 & 0.47 \\
& v3 & 55.14 & 0.73 & 0.58 & 0.55 \\
& v4 & \textbf{85.54} & \textbf{0.85} & \textbf{0.85} & \textbf{0.85} \\
& v5 & 59.02 & 0.58 & 0.58 & 0.57 \\
\hline
\multirow{5}{*}{LLaMA3.2-3B}
& v1 & 34.72 & 0.12 & 0.33 & 0.17 \\
& v2 & 42.18 & 0.74 & 0.46 & 0.38 \\
& v3 & 57.14 & 0.75 & 0.59 & 0.57 \\
& v4 & \textbf{86.59} & \textbf{0.86} & \textbf{0.86} & \textbf{0.86} \\
& v5 & 35.47 & 0.60 & 0.39 & 0.30 \\
\hline
\multirow{5}{*}{Gemma3-4B}
& v1 & 36.66 & 0.12 & 0.33 & 0.18 \\
& v2 & 42.77 & 0.71 & 0.46 & 0.38 \\
& v3 & 55.14 & 0.74 & 0.58 & 0.55 \\
& v4 & \textbf{86.14} & \textbf{0.86} & \textbf{0.86} & \textbf{0.86} \\
& v5 & 28.61 & 0.10 & 0.33 & 0.15 \\
\hline
\end{tabular}
\end{table}
Table~\ref{tab:backbone} presents the performance of the proposed pipeline across different LLM backbones on the test set.

First, all models achieve consistently high performance at the hybrid stage (v4), reaching \textbf{85–86\% accuracy} and \textbf{Macro F1 around 0.85–0.86}. 

Second, this level of performance is close to human annotation accuracy (Table~\ref{tab:annotator_acc}), where annotators achieve approximately \textbf{88–90\%}. This indicates that the proposed system approaches human-level reliability for this task.

Finally, the results remain stable across different LLM backbones, with minimal variation despite differences in model size and architecture. This confirms that the proposed method is largely independent of the underlying LLM and does not rely on a specific model to achieve strong performance.

\begin{table}[h]
\centering
\caption{Comparison between text-only and proposed pipeline}
\label{tab:text_vs_pipeline}
\renewcommand{\arraystretch}{1.1}
\setlength{\tabcolsep}{5pt}
\small
\begin{tabular}{l l c c c c}
\hline
\textbf{Model} & \textbf{Method} & \textbf{Acc} & \textbf{Prec} & \textbf{Rec} & \textbf{F1} \\
\hline
\multirow{2}{*}{Qwen2.5-7B}
& Audio$\rightarrow$Text & 44.11 & 0.38 & 0.41 & 0.341 \\
& Pipeline & \textbf{85.84} & \textbf{0.85} & \textbf{0.85} & \textbf{0.85} \\
\hline
\multirow{2}{*}{Qwen2.5-14B}
& Audio$\rightarrow$Text & 38.70 & 0.39 & 0.38 & 0.362 \\
& Pipeline & \textbf{85.54} & \textbf{0.85} & \textbf{0.85} & \textbf{0.85} \\
\hline
\multirow{2}{*}{LLaMA3.2-3B}
& Audio$\rightarrow$Text & 40.87 & 0.36 & 0.38 & 0.324 \\
& Pipeline & \textbf{86.59} & \textbf{0.86} & \textbf{0.86} & \textbf{0.86} \\
\hline
\multirow{2}{*}{Gemma3-4B}
& Audio$\rightarrow$Text & 40.69 & 0.41 & 0.40 & 0.383 \\
& Pipeline & \textbf{86.14} & \textbf{0.86} & \textbf{0.86} & \textbf{0.86} \\
\hline
\end{tabular}
\end{table}

Table~\ref{tab:text_vs_pipeline} compares the proposed pipeline with a text-only baseline (Audio $\rightarrow$ Whisper $\rightarrow$ Text $\rightarrow$ LLM).

The results show that the text-only approach achieves significantly lower performance (38–44\% accuracy), while the proposed pipeline consistently reaches over \textbf{85\% accuracy}, demonstrating the importance of incorporating acoustic information beyond textual content.

Overall, the experimental results confirm that the proposed pipeline provides a robust and effective solution for speech emotion recognition.

\section{Conclusion}
In this study, we propose a hybrid speech emotion recognition pipeline that combines machine learning with LLM-based reasoning. Experimental results show that the proposed method significantly outperforms zero-shot and text-only approaches, achieving over \textbf{85\% accuracy} and approaching human-level agreement. The method also demonstrates consistent performance across different data splits and remains stable across multiple LLM backbones, highlighting its model-agnostic nature.

Despite these promising results, challenges remain in handling ambiguous emotional states and limited dataset scale. Future work will focus on expanding the dataset, improving reasoning strategies, and extending the system to real-time telehealth emergency scenarios, as well as scaling to multilingual settings for broader applicability.


\begin{thebibliography}{00}

\bibitem{b1} M. El Ayadi, M. S. Kamel, and F. Karray, ``Survey on speech emotion recognition: features, classification schemes, and databases,'' \textit{Pattern Recognition}, vol. 44, no. 3, pp. 572--587, 2011.

\bibitem{b2} B. Schuller and A. Batliner, \textit{Computational paralinguistics: emotion, affect and personality in speech and language processing}. Wiley, 2014.

\bibitem{b3} S. Amershi, M. Cakmak, W. B. Knox, and T. Kulesza, ``Power to the people: the role of humans in interactive machine learning,'' \textit{AI Magazine}, vol. 35, no. 4, pp. 105--120, 2014.

\bibitem{b4} F. Eyben, M. Wöllmer, and B. Schuller, ``OpenSMILE – the Munich versatile and fast open-source audio feature extractor,'' in \textit{Proc. ACM Multimedia}, 2010.

\bibitem{b5} B. McFee, C. Raffel, D. Liang, D. P. W. Ellis, M. McVicar, E. Battenberg, and O. Nieto, ``librosa: audio and music signal analysis in Python,'' in \textit{Proc. SciPy}, 2015.

\bibitem{b6} F. Haider, S. Pollak, P. Albert, and S. Luz, ``Emotion recognition in low-resource settings: an evaluation of automatic feature selection methods,'' \textit{Computer Speech \& Language}, vol. 65, 2020.

\bibitem{b7} S. Latif, R. Rana, S. Khalifa, R. Jurdak, J. Qadir, and B. W. Schuller, ``Deep representation learning in speech processing: challenges, recent advances, and future trends,'' \textit{IEEE Transactions on Affective Computing}, 2021.


\bibitem{b8} A. Holzinger, ``Interactive machine learning for health informatics: when do we need the human-in-the-loop?,'' \textit{Brain Informatics}, vol. 3, pp. 119--131, 2016.

\bibitem{b9} A. Radford, J. W. Kim, T. Xu, G. Brockman, C. McLeavey, and I. Sutskever, ``Robust speech recognition via large-scale weak supervision,'' 2022, arXiv:2212.04356.

\bibitem{b11} G. Trigeorgis, F. Ringeval, R. Brueckner, E. Marchi, M. A. Nicolaou, B. Schuller, and S. Zafeiriou, ``Adieu features? end-to-end speech emotion recognition using a deep convolutional recurrent network,'' in \textit{Proc. ICASSP}, 2016.

\bibitem{b12} M. Neumann and N. T. Vu, ``Attentive convolutional neural network based speech emotion recognition: a study on the impact of input features, signal length, and acted speech,'' in \textit{Proc. INTERSPEECH}, 2017.

\bibitem{b13} P. Tzirakis, G. Trigeorgis, M. A. Nicolaou, B. Schuller, and S. Zafeiriou, ``End-to-end multimodal emotion recognition using deep neural networks,'' \textit{IEEE Journal of Selected Topics in Signal Processing}, 2017.

\bibitem{b14} A. Baevski, H. Zhou, A. Mohamed, and M. Auli, ``wav2vec 2.0: a framework for self-supervised learning of speech representations,'' in \textit{Proc. NeurIPS}, 2020.

\bibitem{b15} N. D. Quang-Anh, M. H. Ha, Q. C. Nguyen, and T. H. Nguyen, ``VNEMOS: Vietnamese speech emotion inference using deep neural networks,'' in \textit{Proc. ICDV}, Hanoi, Vietnam, 2024.

\end{thebibliography}
\end{document}